\documentclass[12pt]{article}
\usepackage{svg}
\usepackage{amsmath}
\usepackage{amssymb}
\usepackage{blindtext}
\usepackage{caption}
\usepackage{epigraph}
\usepackage{lipsum}
\usepackage{authblk}
\usepackage[labelfont=bf]{caption}
\usepackage{subcaption}
\usepackage{graphicx}
\usepackage{parskip}
\usepackage{csquotes}
\graphicspath{ {./figures/} }
\usepackage{url}

\usepackage{geometry}
\usepackage{setspace}
\usepackage[
  backend=biber,
  style=apa,
  sortcites=true,
  sorting=nyt,
  natbib=true  
]{biblatex}

\DeclareLanguageMapping{american}{american-apa}

\title{The Book of Life approach:\\Enabling richness and scale for life course research\thanks{\scriptsize{We would like to thank the following people for helpful conversations: Rafael Batista, Seth Berke, Nicolo Cavali, Emily Cantrell, Gareth Cook, Tom Emery, Per Engzell, Arun Frey, Javier Garcia-Bernardo, Tom Griffiths, Flavio Hafner, Chris Hays, Sayash Kapoor, Angela Li, Anders Weile Larsen, Malte Luken, Alessandra Rister Portinari Maranca, Arvind Narayanan, Juan Carlos Perdomo, Stephen Rabanser, Charles Rahal, Hanzhang (Tommy) Ren, Ksenia Sokolova, Nicol\'as Soler, Lisa Sivak, Brandon Stewart, Gert Stulp, Varun Satish, Keyon Vafa, Veniamin Veselovsky, and Jiani Yan. This research was supported by grants from Princeton Precision Health, the Princeton AI Lab, and the Princeton Catalysis Initiative. All results presented here are calculated from non-public registry data from Centraal Bureau voor de Statistiek (CBS), accessed through the Remote Access environment. CBS was not involved in the calculation of any of the results presented here. While the data are not publicly available, academic institutions can apply for access to the Remote Access environment through the CBS. The underlying data cannot be shared outside of the Remote Access environment as it consists of individual-level, privacy sensitive data. To access the data, an institutional license is required for access to the registry.}}
}

\author[1,2,3]{Mark Verhagen}
\author[1,4]{Benedikt Stroebl}
\author[5,6]{Tiffany Liu}
\author[1,4]{Lydia T. Liu}
\author[1,5,6]{Matthew J. Salganik}

\affil[1]{\small{Center for Information Technology Policy, Princeton University}} 
\affil[2]{\footnotesize{Leverhulme Centre for Demographic Science, Oxford University}}
\affil[3]{\small{Amsterdam Health and Technology Institute}}
\affil[4]{\small{Department of Computer Science, Princeton University}}
\affil[5]{\small{Department of Sociology, Princeton University}}
\affil[6]{\small{Office of Population Research, Princeton University}}

\date{\today}

\begin{document}
\pagenumbering{gobble}

\maketitle

\begin{abstract}
For over a century, life course researchers have faced a choice between two dominant methodological approaches: qualitative methods that analyze rich data but are constrained to small samples, and quantitative survey-based methods that study larger populations but sacrifice data richness for scale. Two recent technological developments now enable us to imagine a hybrid approach that combines some of the depth of the qualitative approach with the scale of quantitative methods. The first development is the steady rise of “complex log data,” behavioral data that is logged for purposes other than research but that can be repurposed to construct rich accounts of people’s lives. The second is the emergence of large language models (LLMs) with exceptional pattern recognition capabilities on plain text. In this paper, we take a necessary step toward creating this hybrid approach by developing a flexible procedure to transform complex log data into a textual representation of an individual’s life trajectory across multiple domains, over time, and in context. We call this data representation a “book of life.” We illustrate the feasibility of our approach by writing over 100 million books of life covering many different facets of life, over time and placed in social context using Dutch population-scale registry data. We open source the book of life toolkit (BOLT), and invite the research community to explore the many potential applications of this approach.
\end{abstract}

\newpage
\newpage

\pagenumbering{arabic}

\section{Introduction}

In 1918 W.I. Thomas and Florian Znaniecki published \emph{The Polish Peasant in Europe and America}, which sparked lasting interdisciplinary interest in life course research~\citep{bulmer1986chicago, elder_emergence_2003}. Life course scholars examine how individual life trajectories unfold over time, across domains (education, work, health), and in a broader context (family, community, society)~\citep{bernardi_life_2019, bronfenbreener_toward_1977, elder_emergence_2003}. Unfortunately, despite the enduring appeal of the life course perspective and substantial effort over the past 100 years, life course research remains difficult for many reasons. These challenges limit our scientific understanding of human development and our ability to design policies that effectively support individuals throughout their lives.

Life course research methods can be roughly grouped into two distinct approaches: quantitative and qualitative~\citep{giele_methods_1998}. The first approach prioritizes studying many people using quantitative methods, but at the cost of analyzing only limited information about each person and their context. The second approach prioritizes richer information, but at the cost of studying fewer people. Each approach has its own strengths and weaknesses~\citep{giele_methods_1998}. We find this methodological pluralism attractive because it increases the ability of researchers to craft a research design appropriate for their research question.  It also increases the space of questions that can be studied and the ways to answer them.

In this paper we propose a new hybrid that we call the ``book of life" approach. Like the quantitative approach, it enables the study of many people.  Further, it takes inspiration from the rich and flexible information that can be used in the qualitative approach. For example, Volume 3 of \emph{The Polish Peasant} included 300 pages on the life history of Wladek Wiszniewski, a Polish immigrant in Chicago, containing information ranging from his upbringing in a peasant family to his later life in Chicago, covering aspects such as work, marriage and family life ~\citep{thomas_polish_1918,plummer_herbert_1990}. It is hard to imagine having a similar amount of richness available in a survey-based quantitative life course analysis. With the book of life approach we strive to make progress in marrying scale-based methods with more information rich data. Our approach is not perfect for every research question and has strengths and weaknesses, just like existing approaches. Some of some of these strengths and weaknesses are described in this paper and some remain to be discovered. 

One rough way to conceptualize the book of life approach is that it transforms data researchers typically associate with quantitative analysis into a form that researchers typically associate with qualitative analysis. More specifically, the book of life approach involves combining different kinds of data into a single piece of text, much like the biography of Wladek.  We term the product of this integration process a ``book of life," drawing inspiration from~\citet{dunn_record_1946}, who envisioned each person’s life as representable in a comprehensive book spanning from birth to death. The metaphorical pages of these books were scattered across numerous databases, recorded in inconsistent formats, or sometimes not recorded at all. Dunn imagined systematically collecting and integrating these different sources of information.

We believe that the time is right to return to Dunn's vision because of two developments. First, in the digital age, there has been a gradual increase in the availability of what we call ``complex log data'', which have different characteristics than data that researchers have used previously~\citep{salganik_bit_2017}. Second, the last few years have seen the explosive growth of large language models (LLMs) that have different characteristics from the statistical and machine learning models that researchers have used previously~\citep{bommasani2021opportunities}. We expect both these trends to continue, and thus we expect the opportunities provided by the book of life approach to grow.

The book of life approach brings many strengths, as we will show in this paper. But it also has many weaknesses, and we think two are important to highlight at the beginning.  The first weakness is that it draws on logged data about behavior. This kind of logged behavioral data is not available in all contexts of interest to researchers. It will likely not include illegal behavior, nor will it include information about individual's subjective experiences. That said, we believe complex log data is well suited to many important questions. The second weakness is that our approach, like many new approaches, currently has limited evidence of effectiveness.  Therefore, we consider the ideas in the paper to be a proof of concept that remains to be fully explored. 

\begin{figure}[!t]
    \centering
    \includegraphics[width=\linewidth]{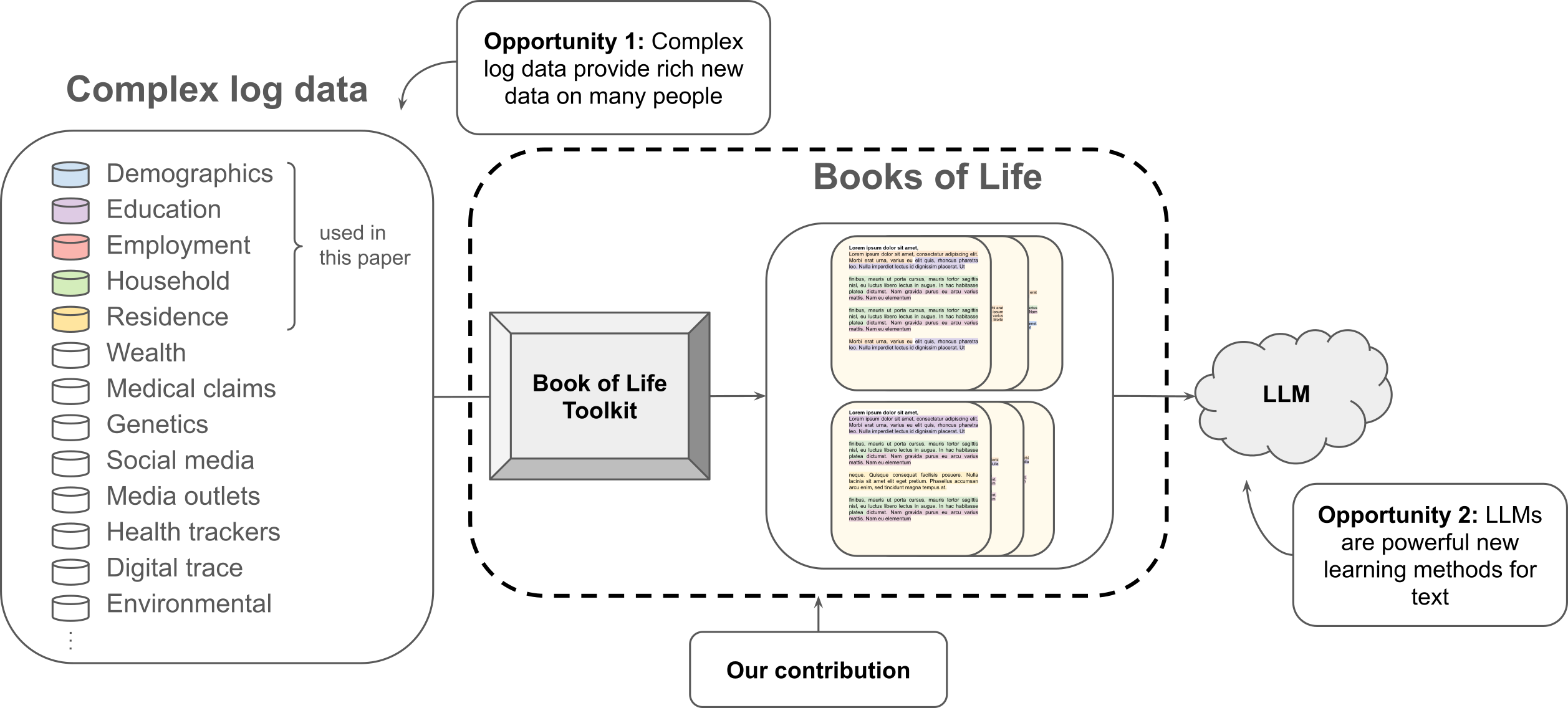}
    \caption{Illustration of the Book of Life approach for life course research, combining richness and scale. Complex log data on the left cover rich information on people's lives but scattered across many files that were created for purposes other than research, spanning different units of analysis, temporal resolution and relational linkages. Books of life express the information for a single individual in plain text. The Book of Life Toolkit (BOLT) in the middle allows mapping complex log data to books of life. An example downstream use case of the books of life approach using LLMs is described in \citet{satish2025}.}
    \label{fig: fig1}
\end{figure}

The book of life approach is schematically illustrated in Figure \ref{fig: fig1}, showing how various types of complex log data are represented in a unifying way as a book of life through the Books of Life Toolkit (BOLT). We developed a first attempt at this pipeline and open source version of BOLT  as part of the Predicting Fertility Data Challenge (PreFer)~\citep{Sivak2024}. PreFer is an international competition to predict fertility in the Netherlands using a subset of one of the world’s richest collections of social datasets: the Dutch population registry. The Dutch population registry is a complex system of heterogeneous datasets~\citep{bakker_system_2014-1} that are created by different government agencies for different legally-mandated purposes. This means the datasets describe different entities, at mixed units of analysis, and with mixed temporal resolution (as will be described in more detail in Section 3). It is difficult to fully capture the richness of the data using standard quantitative methods, positioning the book of life approach as a particularly well-suited starting point.

For PreFer, our task could be divided into two distinct parts: 1) Can we actually construct books of life? and 2) Can we predict fertility based on books of life? In this paper, we focus on constructing books of life in a manner tailored to the specific use case of PreFer, while also supporting a broad range of other applications. In our companion paper~\citep{satish2025} we illustrate the specific downstream use case of PreFer. Concretely, this meant that we generated over 100 million approximately 3-page books of life for 4 million people. There are no fundamental barriers to creating longer books about more people, but this would require additional engineering effort and computational resources as described in more detail below.

We do not think the book of life approach will solve all the limitations or retain all of the benefits of existing approaches. Nor do we think our current implementation within the sandbox of PreFer covers all possible use cases. We are also mindful that we take the quantitative school as our starting point and from there try to address known problems around representing people's life in a rich way. However, we believe that the book of life approach offers a potentially fruitful way to resolve a long-standing tension in quantitative life course research by bridging richer data with methods developed for scale, a goal that has remained elusive for decades.

The remainder of this paper proceeds as follows. Section 2 discusses prior work that serves as the foundation for the book of life approach. Section 3 contextualizes our work within the sandbox of PreFer. Section 4 presents our approach to writing the books of life and describes the Book of Life Toolkit (BOLT), open source software that scalably and flexibly transforms complex log data into books of life. Section 5 concludes and outlines promising directions for future research.

\section{Prior Work}

We can imagine two broad ways to try to marry information richness and scale. One approach would take the qualitative tradition as a starting point and explore ways to add scale. One example of this approach is the ``American Voices project'' that facilitate the mass collection of in-depth semi-structured interviews~\citep{edin_listening_2024}. We instead take the quantitative tradition as a starting point and aim to add richness in the form of more information to it.  In this section, we review known challenges with existing quantitative methods and provide a simple example of a life trajectory that is difficult to capture in its full richness within the quantitative tradition. We then describe how the combination of two developments make new approaches to studying the life course possible: 1) the emergence of complex log data and 2) advances in methods in the field of Artificial Intelligence (AI). We then discuss recent work we most directly build upon. 

\subsection*{The Tension in Quantitative Life Course Analysis}
To illustrate the tension in life course analysis, we introduce two distinct but related components of life course research. The first is a way to represent information that captures the lives of people in a form that can be analyzed, which we call the \emph{data representation}. The second is a way to learn from this representation, which we call the \emph{learning method}. Thomas and Znaniecki used text as their data representation and deep reading as their learning method. This allows for rich detail per case, combined with a flexible learning method---their minds---that allows them to use all their contextual understanding of social life in learning from the data.

By contrast, the dominant data representation for analysis in quantitative life course research is a rectangle of numbers. There is one row for each person, one column for each variable, and a numeric, non-missing value in each cell. The standard learning method is a statistical model that either explicates how each column relates to one another, like a Generalized Linear Model (GLM) or more flexible methods that find such patterns themselves across the columns, like traditional Machine Learning (ML) models~\citep{hastie_elements_2009}.

Unfortunately, representing something as complex as the life trajectories of many people in a rectangle of numbers is challenging. Consider the following example, which is much simpler than Wladek’s 300 page life history in \emph{The Polish Peasant}. 

\begin{displayquote}
``From January 1, 2000 to June 12, 2019 James lived in the municipality of Amsterdam. From June 13, 2019 to December 1, 2019 James lived in the municipality of Leeuwarden. From December 2, 2019 onward James lived in the municipality of Amsterdam.''
\end{displayquote}

This information, when represented in text, seems incredibly simple. A child can read, write, and analyze paragraphs like this. Someone with knowledge of the Netherlands could place this information in a broader context. For example, one might note James grew up in the capital of The Netherlands and then moved to a smaller city in the north of the country when he was 19, but moved back after only a couple months. A qualitative scholar could read and process this simple paragraph and might already identify complex patterns within.

Conversely, turning this information into one row in a rectangle of numbers to analyze with quantitative methods would be more difficult. This requires all information on James to be included as a number \emph{in a single row} of a rectangle of numbers. Recording every residential spell as a separate row with a start- and end-date and a municipality does not suffice because the information about James is spread across many rows.

Next, consider the following minor additions, which adds some richness about James's residential context:

\begin{displayquote}
``From January 1, 2000 to June 12, 2019 James lived \textbf{with his parents} in the municipality of Amsterdam. From June 13, 2019 to December 1, 2019 James lived \textbf{alone} in the municipality of Leeuwarden. From December 2, 2019 onwards James again lived in the municipality of Amsterdam \textbf{with Lisa}. \textbf{Lisa had been living in Amsterdam her entire life.}''
\end{displayquote}

This richer text can still be read and analyzed by a child without problem. Including all of this information into a rectangle of numbers, however, is very complicated. We would need to incorporate a dynamic network perspective that captures James’ changing housemates over time, while also embedding Lisa’s life history within the same row.

Now imagine that we have information like this for millions of people and including additional life domains---like employment or educational histories. Transforming millions of these textual narratives into a single rectangle of numbers requires many arbitrary decisions and would incur considerable information loss.  In other words, lots of data from the life course would no longer be in the rectangle of numbers; it would be possible to go from the stories to the rectangle of numbers, but not the other way around.

So why has it remained so hard to resolve the tension between richness and scale? The reason is that data representation and learning methods are closely intertwined and changing one requires changing the other. As long as a GLM or traditional ML are the predominant learning methods, the rectangle seems natural---even inevitable---as the data representation. Similarly, as long as data on social life is predominantly collected and stored in rectangles of numbers, there is little incentive to explore new learning methods beyond the tabular paradigm. This brings us to two opportunities that, together, motivate us to revisit this decades-old tension.

\subsection*{New data: complex log data}
Waldek’s life history and the survey data more commonly used in quantitative analysis are similar in an important way: both were collected for the purpose of research and with a particular learning method in mind. When data has to be actively collected, it makes sense to align them with existing learning methods. Today, large swathes of information about social life are collected and stored in ways that were never meant for research \citep{salganik_bit_2017}. These data often have similar characteristics that lead us to call these types of data \emph{complex log data}.

Complex log data typically represent time stamped information on events, as they are digitally logged. They are typically stored in complex formats, spanning multiple units of analysis, varying levels of temporal granularity, and dispersed across numerous files~\citep{kashyap2023digital, mcfarland2016sociology}. National registries---the type of complex log data we use in this paper---are a case in point~\citep{penner2019using, bakker_system_2014-1}. These contain records on many life domains, like where someone lives and with whom, where they work, and what type of healthcare they consume. Some might be logged yearly and at the household level, whereas others might be logged daily at the individual level, or monthly at the employer level. In terms of the information they contain, complex log data more closely resemble the level of detail one would find in Wladek’s life history.

When defining complex log data in this general way, a household survey or in-depth interview can be viewed as a very specific type of complex log data. However, the bulk of complex log data does not resemble either. The broad availability of complex log data increases the cost of using a rectangle of numbers as a data representation. We now have more information on social life that is readily available for analysis, but representing this data to align with existing quantitative learning methods runs into the challenges illustrated earlier. Phrased differently complex log data provide the raw material for newer, more information rich types of data representation, but without new learning methods we risk foregoing this potential by forcing them into a rectangle of numbers.

\subsection*{New methods: LLMs}
This brings us to the second opportunity motivating our work. The challenge of developing information rich data representations to learn from is well-known in AI~\citep{bengio2013representation, Lecun2015}. Early research in natural language processing and computer vision used human knowledge to craft sophisticated numerical representations of raw images and text for various computational tasks~\citep{marr1982representation, Jurafsky2009}, but these were eventually outperformed by modern deep learning methods that learn statistics directly from the underlying information~\citep{Lecun2015, krizhevsky2012imagenet}. Since then, these methods have become increasingly powerful, centered around the core idea to learn features from the data rather than engineering them based on the researcher's own intuition~\citep{Sutton2019}.

In natural language processing, this paradigm shift enabled models to learn rich, non-linear dependencies and complex contextual relationships without relying on rigid domain-specific assumptions or manual feature engineering. The result has been a wave of transformative progress, culminating in the LLMs we see today that are able to identify rich patterns directly from sequences of text. Learning approaches that are applied to the raw information and rest on minimal feature engineering are now standard in machine learning research involving text ~\citep{devlin2019bert, radford2018improving, radford2019language, brown2020language}, images~\citep{chen2020simple}, and audio ~\citep{baevski2020wav2vec}.These methodological developments suggest that switching from a rectangle of numbers to other data representations might be fruitful for life course research.

\subsection*{Recent work}
Combining the challenge of using rectangles of numbers to study the life course and the opportunity provided by complex log data and LLMs culminates in our proposed book of life approach. We are not the first to harness the opportunities of complex log data or LLMs for life course research. However, our approach combines different components of past work together in a novel way. We believe this prior work falls into three broad strands.

A first strand of related work retains the rectangle of numbers as the data representation and applies modern AI models directly to these rectangles of numbers~\citep{hollmann2025accurate}. This work shows that these new learning methods are competitive with and sometimes even better than traditional ML methods for tabular data.

A second strand of related work breaks away from the rectangle of numbers and uses these new methods, but does not transform the underlying data to text. For example, \citet{vafa2022career} use a large corpus of online resumes and turn them into sequences of career trajectories, \citet{Savcisens2024} use Danish administrative data covering information on many different life domains to construct temporal life sequences, and \citet{vecgaile_predicting_2025} use pension insurance data to construct life sequences. However, all three choose to represent their data in a bespoke language, rather than text.  These works build on earlier sequence models (e.g., single, cohabiting, married) in the social sciences~\citep{abbott1995sequence, liao2022sequence}. 

The third strand of work applies these new methods to plain text. For example, \cite{Gardner2024} write out rectangles of numbers as plain text and \citet{athey2024labor} study career sequences like \citet{vafa2022career}, but write out careers in plain text. This allows the researchers to use pre-trained LLMs that have already learned rich context on social life as a starting point.

Our work takes elements of this prior work and puts them together in a different way. Namely, we use a non-tabular data representation ~\citep{vafa2022career, Savcisens2024, athey2024labor} that span multiple domains of life~\citep{Savcisens2024}, using plain text~\citep{Gardner2024, athey2024labor} and accounting for the interdependence between people. In doing so, we marry the potential of text as a data representation with the ability of LLMs to learn from such representations. 

\section{The sand-boxed environment of PreFer}
\label{sec:sandbox_prefer}

We tested the feasibility of the book of life approach as part of the international Predicting Fertility challenge (PreFer). The context of the PreFer sandbox shaped many of our decisions on how to write books of life. Below we outline three key contextual dimensions: 1) the goal of PreFer; 2) its computational constraints; and 3) the Dutch registry.

\paragraph{The goal of PreFer.} PreFer’s aim was simple: maximize predictive accuracy on held-out data~\citep{Sivak2024}. This prediction focus diverges from much of life course research, which prioritizes mechanisms and causality~\citep{shmueli2010explain, verhagen2022pragmatist}. Although we targeted predictive accuracy for the challenge, we believe the book of life approach can also potentially support other kinds of estimands~\citep{lundberg_what_2021, ludwig2025large}.

\paragraph{Computational constraints.} All work had to occur inside Statistics Netherlands’ two secure computing environments: a shared virtual desktop server (named the \emph{CBS RA}) and a high-performance computing facility (named \emph{OSSC}). For logistical reasons, we developed books of life solely on the CBS RA (4 CPUs, 48GB RAM (boostable to 128GB) and 100GB of local storage). This meant that our computing environment for generating books of life was broadly similar to those available to many social science researchers. Also, we were able to construct the software and subsequent books in less than 4 months; we only had access from July 4 to October 28, 2024.

Computing constraints extended to the downstream use case of our books of life, too \citep{satish2025}. Larger books were more computationally intensive to analyze, so we limited our books to approximately 1,000 tokens---roughly equivalent to three pages of text. This is modest in comparison to the 300 pages of text on Wladek in \emph{The Polish Peasant}. However, if we had greater computational resources available for downstream use, we could have easily constructed longer books of life; we were nowhere near exhausting the data available to us.

\paragraph{The Dutch registry} PreFer drew on a subset of the data in the Dutch registry, which can be understood as an example of complex log data. A key difference between complex log data and survey data is that complex log data is not neatly structured around an individual; important information about a person is spread across many rows and files. As discussed above, this makes it difficult to analyze complex log data using standard methods without running into the challenges introduced by creating a rectangle of numbers.

\begin{figure}[!t]
    \centering
    \includegraphics[width=\linewidth]{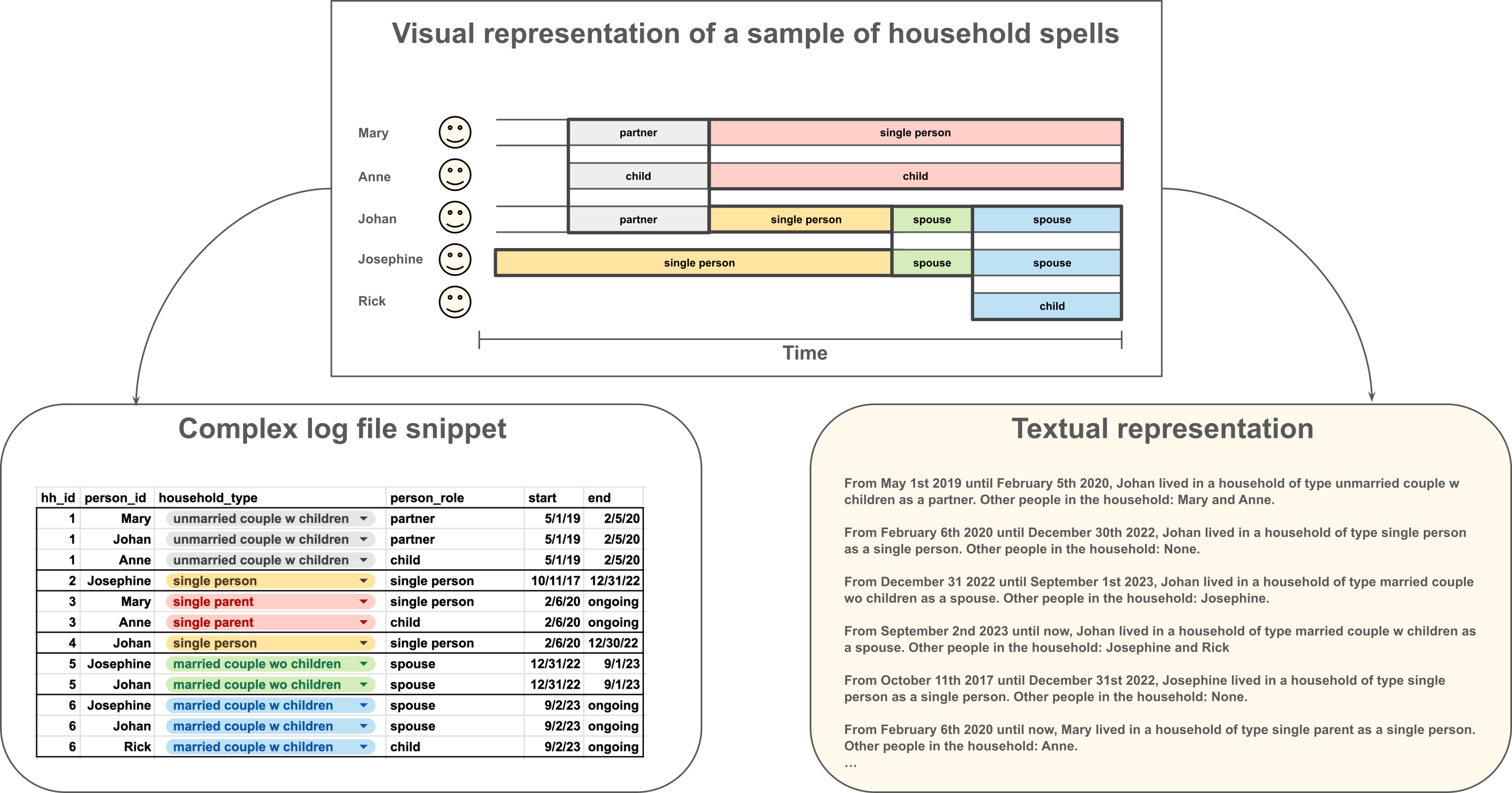}
    \caption{A visual representation of the type of information enclosed in the household log (\emph{Huishoudensbus}) (top), the relevant snippet from the complex log data (bottom left) and the lossless textual representation of this information (bottom right).}
    \label{fig: fig2}
\end{figure}

We illustrate this through one such complex log file in the registry: the household log (\emph{Huishoudensbus}) that records the exact start and end date of every household spell for every individual in the Netherlands. The top of Figure ~\ref{fig: fig2} shows the household spells of a few people organized around a single person: Johan.  Johan lived with his civil partner, Mary, and child, Anne. However, Johan and Mary split up, and Johan moved out. Afterwards, Johan lived alone for a while before marrying and moving in with a new spouse, Josephine. The newlyweds had a child the next year, Rick. All the while, Mary and Anne kept living together. The relevant snippet of information from the household log (\emph{Huishoudensbus}) is shown on the left. On the right, we express this same information as text. Note that these three representations---visual, spells, and textual---all have exactly the same information; it is possible to move in a lossless way between them. 

Now try to imagine about 100 million of these household spells covering the lives of about 20 million people over more than 30 years.
Such logs undoubtedly contain information about life course processes, such as fertility. Unfortunately, learning from this data using standard statistical methods requires the problematic transformation to the rectangle of numbers.

A clever researcher could come up with a reasonable way to transform this file for a specific research question, but the challenges become much more difficult when other logs are also included.  For example, during PreFer we also had access to many other files, including: 1) the employment log (\emph{Spolisbus}), which has one record for every month for every working person for every employment contract; 2) the highest education log (\emph{Hoogsteopltab}), which records the highest educational attainment of each person at the end of every year; 3) a demographic file (\emph{Persoontab}) that records information like sex at birth, birth year and country of birth for every person and their parents.  See Table \ref{table:prefer_tables} for an overview of the files we used, and see~\citet{Sivak2024} for all the files available in PreFer.

\begin{table}[!t]
\centering
\begin{tabular}{|l|p{2.25cm}|p{1.85cm}|p{1.85cm}|p{1.75cm}|p{3cm}|}
\hline
\textbf{File} & \textbf{Content} & \textbf{Unit of analysis} & \textbf{Temporal resolution} & \textbf{Coverage} & \textbf{\# records} \\
\hline
\emph{Spolisbus} & Employment records & Person-Employer-Month & Monthly & Since 2010 & $\sim$1 billion \\
\emph{Huishoudensbus} & Households spells & Person-Household & Daily & Since 1995 & $\sim$200 million \\
\emph{Persoontab} & Demographic information at birth & Person & Monthly & Since 2010 & $\sim$20 million \\
\emph{Hoogsteopltab} & Educational records & Person-Year & Yearly & Since 1999 & $\sim$350 million \\
\emph{Objectbus} & Residential records & Person-Object & Daily & Since 1995 & $\sim$75 million \\
\hline
\end{tabular}
\caption{Complex log data used within PreFer, their content, unit of analysis, temporal resolution, coverage, and number of records. For a detailed list, see \citet{Sivak2024}.}
\label{table:prefer_tables}
\end{table}

These files also raise a fundamental question: if we want to integrate lots of information about a single person into a book of life, how much information should we include about the lives of other people? For example, if we were writing the book of Johan, how much information should be included about Mary, Anne, Josephine, or Rick? What about Johan's parents or co-workers?  The network of potentially relevant individuals expands recursively, forming a web of linked lives~\citep{carr_linked_2018}. We view the ability to include information on lived lives as a key advantage of the book of life approach, but this also unlocks the possibility for infinite regress. We return to this point in the Discussion.

Besides the complex log files in the Dutch registry, we also had access to a ``starter pack'' dataset that was prepared by the PreFer organizers~\citep{Sivak2024}. This starter pack was a rectangle of numbers engineered from the underlying PreFer data. We believe the organizers produced it because they wanted to include the PreFer information in a format that was immediately amenable to standard statistical methods. However, this starter pack also illustrates the challenges of turning complex log data into a rectangle of numbers. For example, the household log (\emph{Huishoudensbus}) described above was included by only recording the household type at the end of 2020 and 2019 for each individual; the rest of the information in the household log was effectively lost.

\section{The Book of Life}

There are many possible ways to integrate complex log data into books of life.  Rather than coming up with one single way to do it, we defined a series of operations that can be flexibly applied on many kinds of data sources to support a large amount of potential books of life.

We also develop open source software that makes it possible for other researchers to explore this space.  We call this software the Books of Life Toolkit (BOLT). BOLT is part of a long tradition in life course research of taking data stored in one format and transforming it into another format for analysis~\citep{balan1969computerized,alter1999casting,bhrolchain1985general}. For BOLT, this means transforming complex log data that is recorded at different units of analysis and temporal resolutions into a single, textual representation of an individual's life. 

We designed BOLT to allow the user to both conceptualize and construct different kinds of books of life. As a consequence, BOLT requires the researcher to make three distinct specifications. The first is the \emph{what}, referring to the information they want to include in the books. The second is the \emph{who}, reflecting what other individuals---if any---should be included into the books. The third is the \emph{how}, referring to the specific format, order and stylistic approach that should be used to write the book. A researcher formalizes the what/who/how choices through a “recipe” file, which then configures the BOLT software to create books accordingly. 

During PreFer, we generated about 100 million books of life in many different formats---some short and simple, and some complex and quite long (Table~\ref{table:books_of_life}). A full list of BOLT’s various options is available as documentation with the software.\footnote{The code underlying BOLT can be found at \url{https://github.com/MarkDVerhagen/BooksOfLifeToolkit}.} 

\begin{table}[!t]
\centering
\begin{tabular}{|c|p{9cm}|c|}
\hline
\textbf{Book} & \textbf{Content} & \shortstack{\textbf{Approx. books}\\\textbf{per second}} \\
\hline
1 & Sex at birth and year of birth & 200 \\
2 & All demographic information at birth & 200 \\
3 & All demographic information and household spells & 120 \\
4 & All demographic information and employment history & 100 \\
5 & All demographic information and changes to highest education & 150 \\
6 & All demographic information and STORK & 180 \\
7 & All demographic information and household spells with household members & 50 \\
8 & All demographic information and the 5 most recent household spells and the 5 most recent address spells & 40 \\
9 & All demographic information, the 5 most recent household spells with household members and their birth year and sex, the 5 most recent addresses, and all employment history & 20 \\
\hline
\end{tabular}
\caption{An example of nine different books of life and their approximate production speed when generating books for approximately 2 million individuals using the CBS RA computation environment. STORK refers to the individual fertility prediction from a model trained on the PreFer starter set.}
\label{table:books_of_life}
\end{table}

\subsubsection*{Specifying the ``what"}
\begin{figure}[!t]
    \centering
    \includegraphics[width=\linewidth]{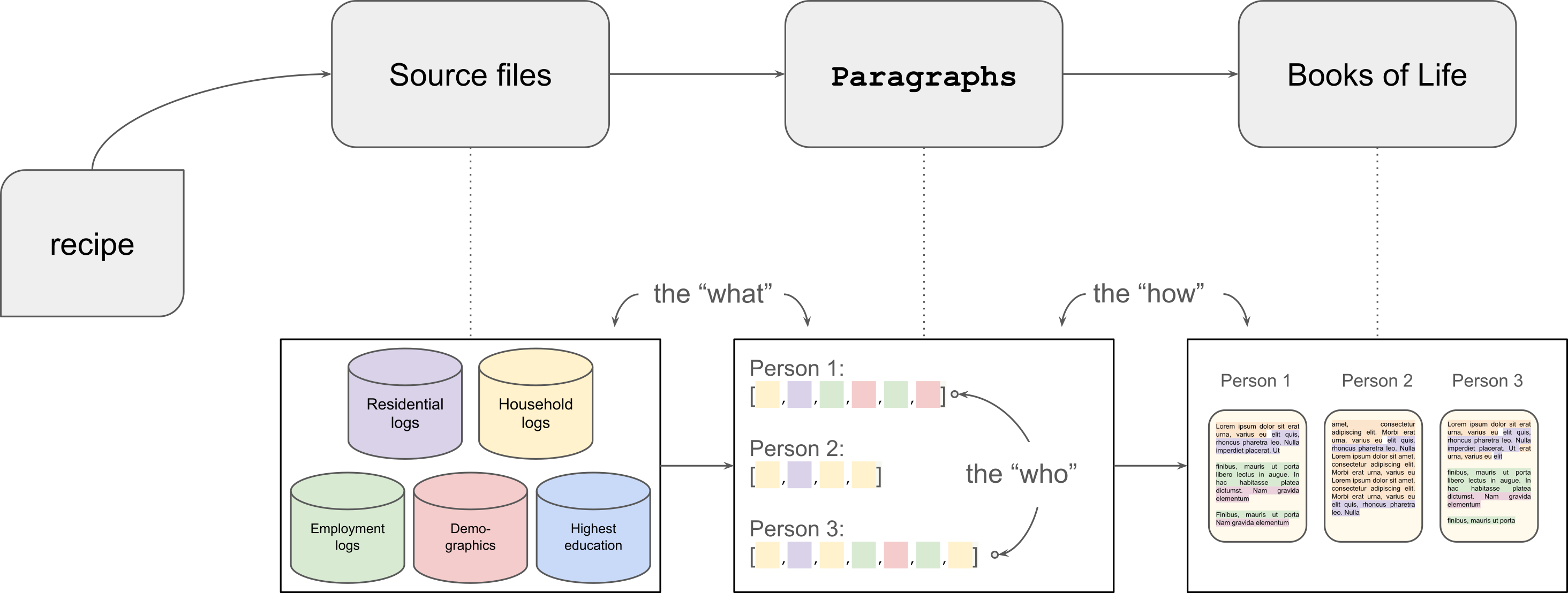}
    \caption{A schematic illustration of the BOLT pipeline. The researcher defines a recipe, which determines what information should be included into the books (the what) and on what people other than the focal person (the who). This information is then collected into a set of paragraphs, which are then written out in a specific way (the how) into a single book of life.}
    \label{fig: fig3}
\end{figure}

Researchers creating a book of life must specify what information, from what files, and over what time frame should be included. In the terminology of BOLT, the candidate information to include in a book is stored in what we call “paragraphs”. For the complex log data we used in PreFer, each individual log entry should be thought of as a possible paragraph. For a rectangle of numbers, a single row should be thought of as a paragraph. This is illustrated schematically on the left of Figure~\ref{fig: fig3}.

To illustrate different ways of specifying the what: the simplest books we created during PreFer included just the focal person’s year of birth and sex at birth (Book 1 in Table~\ref{table:books_of_life}).  We also produced slightly more complex books that included a richer set of demographic information, like the mother's and father’s ages and countries of birth (Book 2). These books closely resemble the information content one would encounter in a survey, as the unit of analysis is the focal person and the information is time invariant.

The complexity of the books can be increased by including information from multiple sources. For example, we produced books that included the same basic demographic information as before but also information about a focal person’s household history (from \emph{Huishoudensbus}) and employment history (from \emph{Spolisbus}) (Book 3 and Book 4, respectively). For Book 4, instead of including every monthly salary slip, we choose to only include “changes” to slips that may indicate a more substantive life event. These could be a promotion (if one’s salary increased a lot), a vacation (if one used many vacation days), or sickness (if one used many sick days).

For Book 5, we similarly choose to only include changes to the highest educational attainment---reflecting the intuition that state changes contain as much information as repeated measures of the states themselves~\citep{bernardi_life_2019}. 

For Book 6 we included a summary metric of the starter pack by the PreFer organizers. Rather than including the entire rectangle, a boosted decision tree was trained to use this data to predict fertility, and then we included the predicted probabilities in each book~\citep{Cantrell2025}. The ability to arbitrarily include summary statistics rather than raw data illustrates some of the flexibility of the books of life approach. 

Across Books 4-6, nothing precluded us from including the entire starter pack or every log into the books of life, but we decided to try to reduce the information to 1) explore the flexibility of a book of life approach and 2) reduce computational demands during downstream use (see Section~\ref{sec:sandbox_prefer} for more on computational constraints).\footnote{To only include the state changes for the employment log (\emph{Spolisbus}) and education log (\emph{Hoogsteopltab}), we implemented some basic preprocessing of the raw log files. In the future, we envision including such functionality as part of BOLT.}

\subsubsection*{Specifying the “who”}
Beyond the information on the focal person, a second way to increase the richness of the book is by adding information on other individuals. For example, in Book 7 we expanded Book 4 to include the individuals that were part of each household of the focal person. The book of life approach and BOLT are flexible enough that we could have included any other set of people, such as all their coworkers---as defined in the employment log (\emph{Spolisbus})---or their neighbors---as defined in the address log (\emph{Objectbus}) as well. In theory, BOLT allows for variables that aren't usually thought of as context variables to be used as well, like adding all other people that have the same birth country as the focal person into their book.

Once the researcher decides who to include, they can decide what to include about these people. At default, the book of life just includes a list of their IDs or, more interestingly, they can choose to write out actual books for them as well.  These “books within books” allow the researcher to flexibly capture extremely rich information about the social context around a person.

\subsubsection*{Specifying the ``how”}
The ``what" and ``who" decisions determine the information in each book. The ``how" decision determines exactly how this information is written. This is illustrated on the right of Figure~\ref{fig: fig3}. Just as the norms for how the information is structured in a physical book evolved over many years based on theory, as well as trial and error, we expect that the norms for how information is written in books of like will evolve. At this point, there are three main options available: filter, order, and style.  But as more is learned about the book of life approach, we expect these options to increase.

\paragraph{Filter: the user can decide to further exclude candidate paragraphs from the book.} For example, the user could choose to only include the last 5 household spells and last 5 employment events (Book 8), rather than all available spells and events.

\paragraph{Order: the user can decide how to order paragraphs in the book.} For example, they could choose to order all paragraphs chronologically. They could also choose to order paragraphs by file and then chronologically.

\paragraph{Style: the user can decide how to write a paragraph into text.} By default, the information in a paragraph is written as a key-value pair where the title of each field of information is followed by a colon and the value for that field. For the most basic Book 1, this would look something like ``\texttt{gbageslacht:\ 1, gbageboortjaar:\ 1990}'', as the field for sex was called “gbageslacht” and coded “1” for male and “2” for female, and the field for birth year was “gbageboortejaar” and coded as a numeric. BOLT supports what we call “parsing dictionaries” that can map both fields and values to more informative values. In the case of Book 1, this meant we wrote the book as “\texttt{Sex at birth:\ Male, Birth year:\ 1990}”.

For certain files we made more elaborate parsing dictionaries. A single paragraph from the household log (\emph{Huishoudensbus}) from Book 7 would normally look something like: ``\texttt{hh\_id:\ 1, person\_id:\ Johan, household\_type:\ 3, person\_role:\ 1, start:\ \\05-01-2019, end:\ 02-05-2020, hh\_person\_ids:\ [Mary, Anne]}”. We added a parsing dictionary that wrote this paragraph out as: “\texttt{From January 5th 2019 until May 2nd 2020, Johan lived in an unmarried couple with children household as a parent.\ The other people in the household were Mary and Anne}”.

Finally, the user can decide to add section headers or other types of styling to each book of life as they see fit.

\subsubsection*{Extended example}

Having broadly explained how the books of life are structured, we will illustrate a BOLT recipe file in more detail for Book 9. We will follow the general structure of the actual recipes but use pseudo code instead of the actual commands that would be used in BOLT.

\begin{figure}[!t]
    \centering
    \includegraphics[width=\linewidth]{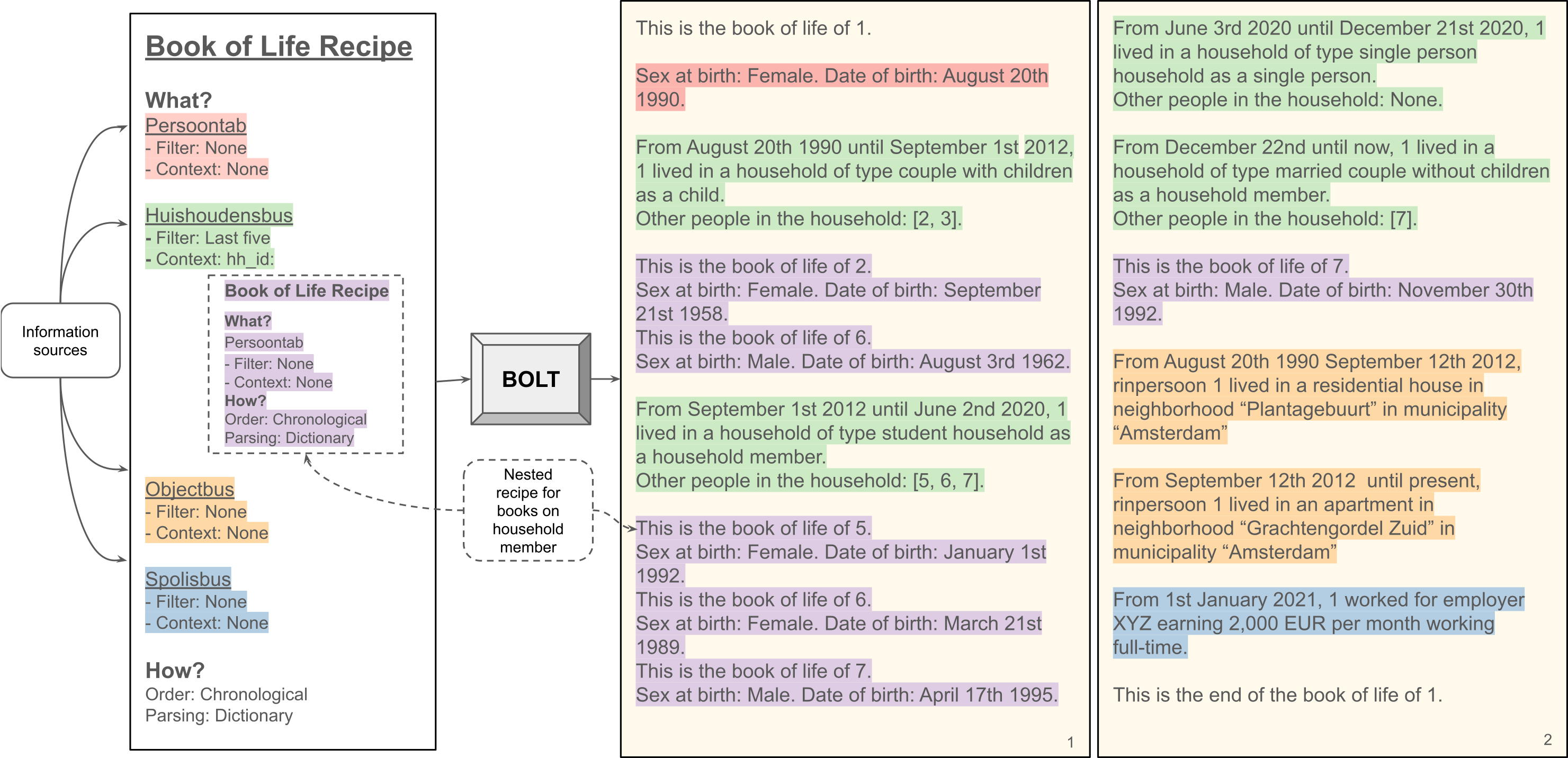}
    \caption{An example recipe (on the left) illustrating what information sources to include and for whom into the book of life and how to write them. On the right, a sample two-page book for an illustrative person “1” is shown. Each information source from the recipe is color-coded in the same way as the associated paragraphs in the book on the right.}
    \label{fig: fig4}
\end{figure}

The recipe is shown on the left side of Figure~\ref{fig: fig4}.  First we define the “what”. This book includes information from the demographic log (\emph{Persoontab}), household log (\emph{Huishoudensbus}), address log (\emph{Objectbus}), and employment log (\emph{Spolisbus}). Next, we define the ``who".  In this case we include all housemates in each spell (taken from the household log (\emph{Huishoudensbus}). We then specify a new recipe, through which we include demographic information on these other household members (indicated in the dashed box). This is an example of a "book within a book". Finally, we define the “how”. For demographic information, address spells and the employment history, we simply include all available paragraphs with no filtering. For the household spells, we only include the final five observed spells. The settings at the end indicate that we want to order all paragraphs chronologically and we want to use parsing dictionaries.

The resultant two-page book of life is provided on the right of Figure 4. The first paragraph is the demographic information and is written in key-value form, but with a parsing dictionary to make the field names and values meaningful. Next, four household spells are included. Although we specified the last five, this person only had four spells in the data as can be seen from the birth date and first household spell. We then include basic demographics on each household member from each spell. Then, we included the address spells (two) and all employment changes (one).

As illustrated in Table~\ref{table:books_of_life}, writing these types of books was feasible with a standard desktop server (the CBS RA) even though the underlying log files all had millions or billions of records (see Table ~\ref{table:prefer_tables}). However, as the richness of the books increased the throughput rates decreased. We want to emphasize that we did not develop BOLT to be computationally optimal, but rather so that it was sufficient to produce books within the limited time we had during PreFer. While we hope the book of life approach will endure, we anticipate that the Book of Life toolkit will soon become outdated as others develop more efficient and general versions that expand its functionalities outside of the sandboxed environment of PreFer.

\section{Discussion and Conclusion}
Our work was motivated by a long-standing desire to combine scale and richness in life course research. Simply put, it is difficult to turn something as rich as the life history of Wladek from \emph{The Polish Peasant}, into a row of numbers. We believe that progress on this challenge has been slow, in part because it requires two things to change: we need data that enable us to construct new, richer data representations of people’s lives, and we need new methods to analyze such representations.

In this paper, we built on two independent developments that allow us to imagine a new way of doing life course research. This approach builds on the widespread availability of complex log data---like the digitized administrative records that we use in this paper---and the ability to analyze sequences of text using LLMs that have already learned rich context about the world based on internet-scale corpora. Thus, text is both a natural and flexible way to represent a life trajectory and a way to tap into the power of LLMs as they exist today and may develop in the future.

That said, the ultimate value---and limitations---of the books of life approach remain to be explored. Our first attempt at producing books of life at scale were based on a single, albeit rich, source of information---the Dutch Registry---and were focused on a particular estimand: predictive performance. There may be many other downstream tasks beyond predictive ability for which books of life could be used. This brings us to three areas that we think are critical avenues for future work.

\subsubsection*{More and different data}
The data used in PreFer is just a small subset of the data in the Dutch registry, and the data in the Dutch registry, albeit considerable, is still a subset of all the data we might want to use to study the life course. A natural extension to our work would be to write longer, richer books, getting closer to the 300 pages about Wladek that were in \emph{The Polish Peasant} over a century ago.

As a first step we are particularly interested in enriching books to include information related to health.  For example, the Dutch registry includes complete medical billing records and complete prescription drug records for over a decade.  Further, Statistics Netherlands regularly conducts surveys on hundreds of thousands of people to collect information on self-reported health (\emph{Gezondheidsmonitor}). In these millions of life-health trajectories, we hope to discover new insights into the social origins of clinical health outcomes.

We also think it is worth including more levels of context into the books~\citep{bronfenbreener_toward_1977}. These could include local newspaper articles or government statistics about the Dutch economy at the city, regional, or national level. We expect that new approaches to ``how" will need to be developed to effectively include these kinds of contextual layers into books of life. 

Finally, we think it makes sense to include non-textual information in books and make them multi-modal, in line with recent developments in LLMs. Just like life course scholars frequently resort to graphic depictions of ideas, people, or contexts, we would similarly envision enriching our books of life with such content. If it is true that a picture is worth a thousand words, multi-modal books should create new research opportunities. 

\subsubsection*{A bigger role for LLMs in creating books of life}
Due to the logistical constraints of PreFer, we developed BOLT in an environment where we could not use LLMs to help write the books of life.  One advantage of this constraint is that it led us to create a structured framework to develop books, where a researcher has full control of what is to be included and how it is to be written.  This parameterizable approach helps us better map the range of possibilities. 

However, it also introduces many design choices into the process. More broadly, the ability to recursively include ``books within books'' means that the amount of information one can include in a book can quickly become large. With this in mind, we see a promising direction in exploring the use of LLMs not only for analyzing books of life, but also for contributing to their construction.

We think the most exciting directions for using LLMs to create books of life are not for LLMs to replace BOLT but to work with the toolkit.  For example, researchers could generate rich books of life using BOLT, and then use an LLM to summarize books depending on the desired downstream use.  We believe that the biggest gains from this LLM-in-the-loop approach could come from making this process end-to-end, whereby the summarization by the LLM is constantly refined based on feedback from the performance of the downstream task. 

Further, one could imagine using LLMs as part of a multi-agent workflow~\citep{guo_large_2024}. For example, one LLM agent would be tasked with writing and running the recipe file to produce a set of books of life, a second LLM agent would then be tasked with using the books of life for some downstream task, and a third LLM would summarize the results.  These three LLM agents would discuss the results, possibly against the relevant scientific literature, to decide what to do next. This multi-agent approach is highly scalable; there could be thousands of agent teams running simultaneously, which could be seen as a step toward what some have called a ``country of geniuses in a datacenter"~\citep{amodei_machines_2024}.  

While we cannot imagine something as grand as ``a country of geniuses" in the near future, for a narrow set of downstream tasks we think it is probably possible to create something that approaches a few dozen aspiring scholars working with 100\% focus for long periods of time.  Thus, rather than a country of geniuses, we think these multi-agent systems could be more like mass collaboration for predictive modeling similar to the Fragile Families Challenge, where teams of researchers contributed models to predict life outcomes from a shared social dataset~\citep{Salganik2020}.  While each team of LLM agents would likely do worse than a team of skilled human researchers, it would be possible to have thousands working in parallel in a way that is not logistically feasible with human teams.  We expect that in some situations, novel approaches might come from this high-throughput exploratory approach to discovery, even if most of the approaches are not very impressive. 

\subsubsection*{Governance}
The book of life approach to life course research relies on the existence of centralized data repositories. However, centralizing personal data, especially in the case of population registries, raises the possibility of misuse~\citep{Seltzer2001}. While our approach benefits from centralized data, we do not necessarily advocate for further centralization of social data to enable methods like ours. Instead, we argue that where large-scale data repositories of personal data already exist---such as in national administrative datasets---they must be governed effectively to balance research access with appropriate safeguards. We believe that such governance mechanisms exist and were in place in the setting in which we worked~\citep{Sivak2024}. Future research could explore the extent that the book of life approach could be done in a federated manner where different information sources are still controlled by different data custodians.

We note that the governance questions would increase dramatically if these techniques were moved outside of the sandboxed setting of academic research into a real-world setting where they inform decisions in high stakes situations.  Such approaches with population registry data are legally prohibited by Dutch law, nor is it possible to export individual-level predictions from the secure Statistics Netherlands computing environment.

To conclude, this paper is about re-imagining a century-old aspiration of social research: to understand human lives across many domains, over time, and in context. Traditionally, scholars attempting to realize this vision have had to sacrifice either richness or scale. Taking advantage of new data (complex log data) and new methods (LLMs), we develop the book of life approach that enables both information richness and scale. The strengths and weaknesses of this approach remain to be discovered by the research community, and we hope that our open source book of life toolkit (BOLT) will help with that exploration.  We are optimistic that much remains to be discovered when a century-old tradition is combined with new data and new methods.

\newpage

\printbibliography

\end{document}